\documentclass[journal,letterpaper]{IEEEtran}
\usepackage{blindtext}
\usepackage{graphicx}
\usepackage{multirow}
\usepackage{color}
\usepackage{fancyhdr}
\usepackage{floatflt}
\usepackage{float}
\usepackage{times}
\usepackage{color,soul}
\usepackage{graphics}
\usepackage{subfigure}
\usepackage[ruled]{algorithm2e}
\usepackage{amsmath, amssymb}
\usepackage{bm}
\usepackage{hyperref}
\usepackage[colorinlistoftodos]{todonotes}
\usepackage{cite}
\usepackage[font=small,skip=2pt]{caption}
\usepackage{amsbsy}

\ifCLASSINFOpdf
\else
\fi
\begin{document}
%
\title{Fast Sequence Component Analysis for Attack Detection in Synchrophasor Networks}

\author{Jordan~Landford,~\IEEEmembership{Student~Member,~IEEE,}
        Rich~Meier,~\IEEEmembership{Student~Member,~IEEE,}
        Richard~Barella,~\IEEEmembership{Student~Member,~IEEE,}
        Xinghui~Zhao,~\IEEEmembership{Member,~IEEE,}
        Eduardo~Cotilla-Sanchez,~\IEEEmembership{Member,~IEEE,} 
        Robert~B.~Bass,~\IEEEmembership{Member,~IEEE,}
        and~Scott~Wallace,~\IEEEmembership{Member,~IEEE}
\thanks{R.~Meier and E.~Cotilla-Sanchez are with the School of Electrical Engineering \& Computer Science, Oregon State University, Corvallis, OR, USA} 
\thanks{R.~Barella, S. Wallace and X. Zhao are with the School of Engineering and Computer Science, Washington State University, Vancouver, WA, USA}
\thanks{R.~B.~Bass and J.~Landford are with the Maseeh College of Engineering and Computer Science, Portland State University, Portland, OR, USA}
}

\maketitle

\begin{abstract}
Modern power systems have begun integrating synchrophasor technologies into part of daily operations. Given the amount of solutions offered and the maturity rate of application development it is not a matter of ``if'' but a matter of ``when'' in regards to these technologies becoming ubiquitous in control centers around the world. While the benefits are numerous, the functionality of operator-level applications can easily be nullified by injection of deceptive data signals disguised as genuine measurements. Such deceptive action is a common precursor to nefarious, often malicious activity. A correlation coefficient characterization and machine learning methodology are proposed to detect and identify injection of spoofed data signals. The proposed method utilizes statistical relationships intrinsic to power system parameters, which are quantified and presented. Several spoofing schemes have been developed to qualitatively and quantitatively demonstrate detection capabilities.
\end{abstract}

\begin{IEEEkeywords}
spoofing, phasor measurement unit, PMU, correlation, event detection, support vector machine.
\end{IEEEkeywords}

\section{Introduction}

Phadke and Thorp's 1988 invention, the phasor measurement unit or PMU, provides power systems operators with near real-time measurements of Steinmetz's current and voltage phasors, thereby permitting improved wide-area monitoring, control and protection of power systems\cite{1178427, Steinmetz1893, 5447627}. Imperative when using PMUs for any of these purposes is to ensure data integrity. Data integrity may be compromised randomly, as data drops or clock drifts, or maliciously via data injection. 

We propose that disruptions to data integrity may be detected by monitoring correlation values between phasor measurements from multiple adjacent PMUs.  In previous work, we show that a matrix of correlation values between a cluster of PMUs can quickly reveal data corruption, particularly data drops and GPS clock drift\cite{UsSusTech14BPAEE}. These kinds of events result in rapid decorrelation between the afflicted PMU and all others, observable as the appearance of a row and column of very low correlation values. The parameters at electrically-close PMUs are normally highly correlated; fluctuations in voltage, phase and frequency are not single-bus behaviours, as adjacent buses will experience similar effects in a well-correlated manner.  

Both data drops and drifts result in very rapid decorrelation. Less likely to be detected would be spoofing attacks whereby vectors of ``typical'' PMU data are somehow injected in place of a PMU's actual output data stream.  Carefully-chosen vectors could be used to disguise an attack that would otherwise alter PMU data measurements, thereby providing the attackers with cover while conducting a malicious attack at a substation.  By leveraging historic archived PMU data, we believe we can characterize the distribution of correlation values during normal operation with enough fidelity to identify many potential spoofing strategies. 

In order to mount a successful attack, vectors of ``typical'' data, which we refer to as ``spoofed'' data, must be injected in place of the actual data stream in order to not raise alarms to the attack.  It is reasonable to assume an attacker could generate a representative vector of positive sequence voltage data, given its propensity to be within $\pm10$\% of 1.0 p.u. Generating a convincing pair of vectors, however, would be less likely, particularly if the parameters are weakly-correlated.  By monitoring correlation values of several parameters between multiple electrically close PMUs, attempts to inject false data may be detected. Data-driven attacks have been suggested in the related literature \cite{jinsub1,jinsub2}. 

In this paper, we propose an approach to detect spoofed signals from PMU data streams by monitoring the change of correlation values between PMUs. The data used in this study are collected by Bonneville Power Administration in their wide-area monitoring system. We first examine intra-PMU and inter-PMU correlations to identify useful features for detecting spoofed signals. These features are then used to train a set of two-class Support Vector Machines (SVMs) for detecting specific types of spoofs. The experimental results on a separate testing data set show that this approach is accurate in detecting the type of spoof the SVM is trained on. To generalize this approach, we use ensemble methods to combine a set of Support Vector Machines described above, so that they can collectively make decisions about a new unknown signal. The results show our spoof detection ensemble is more robust than the individual Support Vector Machines, and demonstrates the generalizability of this approach in terms of identifying spoof signals that have not been previously seen. 

The remainder of the paper is organized as follows. Section~\ref{sec:background} presents background and related research work on spoof attacks and detection. In Section~\ref{sec:methodology}, we analyze the correlation values of our PMU dataset for the purpose of identifying useful features. Section~\ref{sec:results} presents the feasibility of using PMU correlation to detect spoofed signals. In Section~\ref{sec:SVM}, we present the details of our Support Vector Machines, experimental results, as well as the spoof detection ensemble. Finally, in Section~\ref{sec:conclusion} we conclude the paper and present future directions of this research. 

\section{Background} \label{sec:background}

Attacks wherein spoofed data are injected into a SCADA system to disguise an attack have been documented, most notoriously Stuxnet\cite{5772960, 6471059, 5742014}.  Stuxnet was a computer worm designed to be inflicted upon on industrial equipment, specifically Siemens PLCs (programmable logic controllers). The intent of Stuxnet was to physically destroy a specific target, in this case thousands of Iranian uranium centrifuges.  Stuxnet was a sophisticated multi-modal attack for which spoofing was used to mask malicious activities. Specifically, Stuxnet periodically varied the mechanical frequency of the centrifuges while concurrently masking these changes by producing spoofed process control signals. As such, the PLCs would not shut down because they could not observe the abnormal behaviour. 

One lesson of Stuxnet is that physical infrastructure may come under the control of malware. Even isolated industrial systems are vulnerable to physical attack, and sensor spoofing is a means by which such an attack may be masked.  Other critical cyber-physical systems are also susceptible to attack, notably Global Navigation Satellite Systems (GNSS), a susceptibility that has been known for over twelve years\cite{Warner2002, Humphreys2009}. A spoofing detection method for GNSS has been developed by Magiera and Katulski based on measurements of phase delay\cite{6614026}. Similar to our own approach with PMUs, Psiaki \textit{et al.}, use cross-correlation of encrypted signals between two GNSS receivers to detect spoofing of publicly-known signals\cite{6621814}.

Other vulnerable cyber-physical systems include vehicular ad hoc network, and of course, electrical power systems\cite{6117120}. PMUs are becoming critical data sources for multiple power systems functions, providing measurements for state estimators, initiating remedial action schema, and estimating voltage-stability margins\cite{780916}. Threat potential has been demonstrated by Jiang, \textit{et al.}, whereby they maximize the difference between the PMU's receiver GPS clock offset before and after an attack\cite{6451170}. And, Zhang, \textit{et al.} investigated the consequences of an attack on the time stamps of data collected within a smart grid wide-area network\cite{Zhang2011}. Threats to PMUs have been summarized by Shepard, \textit{et al.}\cite{Shepard2012}.

Machine learning techniques have proved to be effective in detecting security attacks in cyber-physical systems~\cite{mitchell2013effect} \cite{amor2004naive}, including smart grid~\cite{kher2012detection}. However, to the best of our knowledge, there is no previous work on detecting spoofed signals injected in real PMU data streams. This paper presents an approach in this direction. 

\section{Methodology} \label{sec:methodology}

We use PMU data from ten electrically-close PMUs from Bonneville Power Administration's 500 kV PMU network.  These data were recorded at 60 frames per second. 

\subsection{Intra-PMU Parameter Correlation}
PMUs measure phasors of line voltages and line currents for all voltages (A, B, C) and currents (A, B, C, N).  From these are derived a number of other parameters, including magnitude and phase angle for the positive, negative and zero sequence voltages and currents; frequency; and rate of change of frequency (ROCOF); among others\cite{1611105}. Some of these parameters show moderate correlation between each other, but most do not.  

We use the Pearson correlation coefficient ($r$) to quantify the degree of correlation between PMU parameters. Intra-PMU parameters that are weakly correlated can be used to detect spoofing attempts using the method described in this paper. Monitoring multiple, poorly correlated PMU parameters makes it more difficult for the attackers to provide convincing spoofed data sets. Table~\ref{tab:PMU_r} shows the mean and standard deviation of correlation values between PMU parameters from a single PMU.  Correlation values were calculated for each time step using a one second wide sliding window.  The mean and standard deviation were then calculated using 59 seconds of data.   

\begin{figure}[ht]
\centering
\includegraphics[width=3.8in]{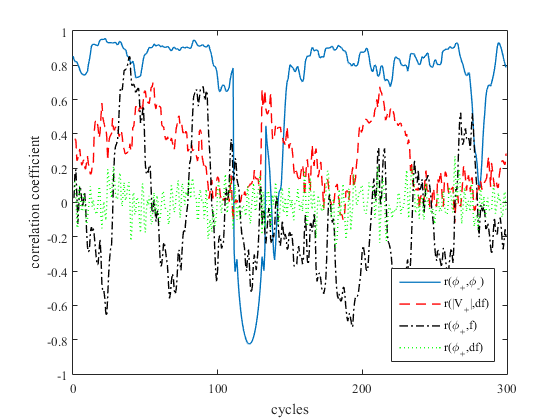} 
\caption{Plot of sliding window intra-PMU correlation values.  High correlation is observed between sequence phase angles, such as $\phi_+$ and $\phi_-$.  Low correlation, with wide deviation is observed between $|V_+|$ and ROCOF as well as between and $\phi_+$ and $f$. Low correlation and low deviation is observed between $\phi_+$ and ROCOF.}
\label{fig:SinglePMUr}
\end{figure}

Most, but not all of the intra-PMU correlation \emph{r} values are near zero, though with wide standard deviations. We observe modest correlation between the phase angles of the sequence components, 0.7 and above. We observe weak correlation, with small standard deviations, between the voltage angles and frequency, as well as between the voltage magnitudes and the rate of change of frequency. The weakest correlation, with very little deviation, we observe between the voltage angles and the rate of change of frequency. Correlations between pairs of intra-PMU parameters as a function of time are illustrated in Figure~\ref{fig:SinglePMUr}. The weakly correlated parameters with narrow standard deviations are the best candidates for use in detecting spoofing attempts, if the correlation of these parameters between adjacent PMUs is strong. 


\begin{table}[h]
\centering
\caption{Mean and standard deviation, $\mu$ $(\sigma)$, of correlation between intra-PMU parameters. $|V_+|$, $|V_-|$ and $|V_0|$ are the positive, negative and zero sequence voltage magnitudes. $\phi_+$, $\phi_-$ and $\phi_0$ are the positive, negative and zero sequence voltage phase angles. $f$ and ROCOF are the system frequency and its rate of change.}
\begin{tabular}{c| c c c c c c c}
& $\phi_+$ & $|V_-|$ & $\phi_-$ & $|V_0|$ & $\phi_0$ & $f$ & ROCOF\\
\hline
$|V_+|$ & -0.06 & -0.02 & 0.07 & 0.07 & -0.03 & 0.32 & 0.17\\
& (0.49) & (0.34) & (0.47) & (0.36) & (0.48) & (0.25) & (0.19)\\
$\phi_+$ & & -0.03 & 0.72 & -0.03 & 0.82 & 0.00 & 0.00\\
& & (0.42) & (0.44) & (0.43) & (0.41) & (0.40) & (0.09)\\
$|V_-|$ & & & -0.06 & 0.06 & 0.03 & -0.11 & 0.01\\
& & & (0.40) & (0.30) & (0.43) & (0.32) & (0.18)\\
$\phi_-$ & & & & 0.11 & 0.67 & 0.04 & -0.01 \\
& & & & (0.40) & (0.42) & (0.37) & (0.12) \\
$|V_0|$ & & & & & 0.04 & -0.02 & 0.01 \\
& & & & & (0.42) & (0.27) & (0.16) \\
$\phi_0$ & & & & & & 0.02 & 0.02\\
& & & & & & (0.39) & (0.11)\\
$f$ & & & & & & & 0.52 \\
& & & & & & & (0.09)
\end{tabular}
\label{tab:PMU_r}
\end{table}

\subsection{Inter-PMU Correlation}
To quantify the degree of correlation between parameters of nearby PMUs, we again use the Pearson correlation. 
For example, given PMUs numbered $1,2\dots ,p$ we develop ${p \choose 2}$ vectors of correlation values between the positive sequence voltage magnitude $R_{ij}(|V_+|)$ for every pair of PMUs $i < j$.
This is repeated for the $|V_-|$, $|V_0|$, $\phi_+$, $\phi_-$, $\phi_0$, $f$ and ROCOF data. These correlation values fluctuate with time, since the correlation is performed using data windows of a fixed length. For this work, we examined windows of 1, 2, 5 and 10 seconds in length.  

We found that the correlation vectors $r(|V_+|)$, $r(\phi_+)$ and $r(f)$ are good candidates for detecting spoofing attacks, as these consistently exhibit moderate to high correlation values over wide ranges of time. The $r(\phi_+)$ correlation values are exceptionally high, near 1.0 under normal circumstances. On the other hand, $r(|V_-|)$, $r(|V_0|)$, $r(\phi_-)$ and $r(\phi_0)$ do not exhibit consistent moderate correlation. 
ROCOF correlation between PMUs is very poor, likely due to the fact that it is the second derivative of the positive sequence phase angle, and hence more susceptible to noise.

\subsection{Modeling a Spoofing Event}
\label{sec:spoof-types}

Our spoofed data set was derived by recording 30 seconds (1800 cycles) of PMU data from one of our ten PMU sites, then playing back these data in a modified form to generate the final 30 seconds of a complete minute. We considered the following spoof playback schemes: 

\begin{itemize}
\item {\em S1: Mirroring} in which the initial 30 seconds of data are played back in reverse to produce the final 30 seconds of data.

\item {\em S2: Polynomial Fit} in which a $3^{rd}$ degree polynomial is fit to the initial 30 seconds of data. This polynomial is combined with a noise profile to generate the final 30 seconds of data.

\item {\em S3: Time Dilation} in which we record a full 60 seconds of data from the spoofed PMU, but resample the final 30 seconds of data so the signal appears stretched over time. For this spoof, we explored various rates of time dilation ranging from very slow (a factor of 2 slower than real-time) to near real time (a factor of 8/7 slower than real-time). Recent studies challenging the security of GPS hardware for PMUs detail the feasibility of this type of spoofing approach \cite{alejandro_gps,sg_gps,security_gps}.
\end{itemize}

Each spoofing approach above guarantees signal continuity for all parameters at the instance spoofing is initiated, and correlation for a brief time after that transition. All ten of the monitored PMU sites are electrically close, showing strong correlation between their frequency measurements. 

\section{Spoof Detection Using PMU Correlation}
\label{sec:results}

\begin{figure}[ht] 
  \centering
    \subfigure[120 cycle (2 seconds) window size]{\includegraphics[width=3.5in]{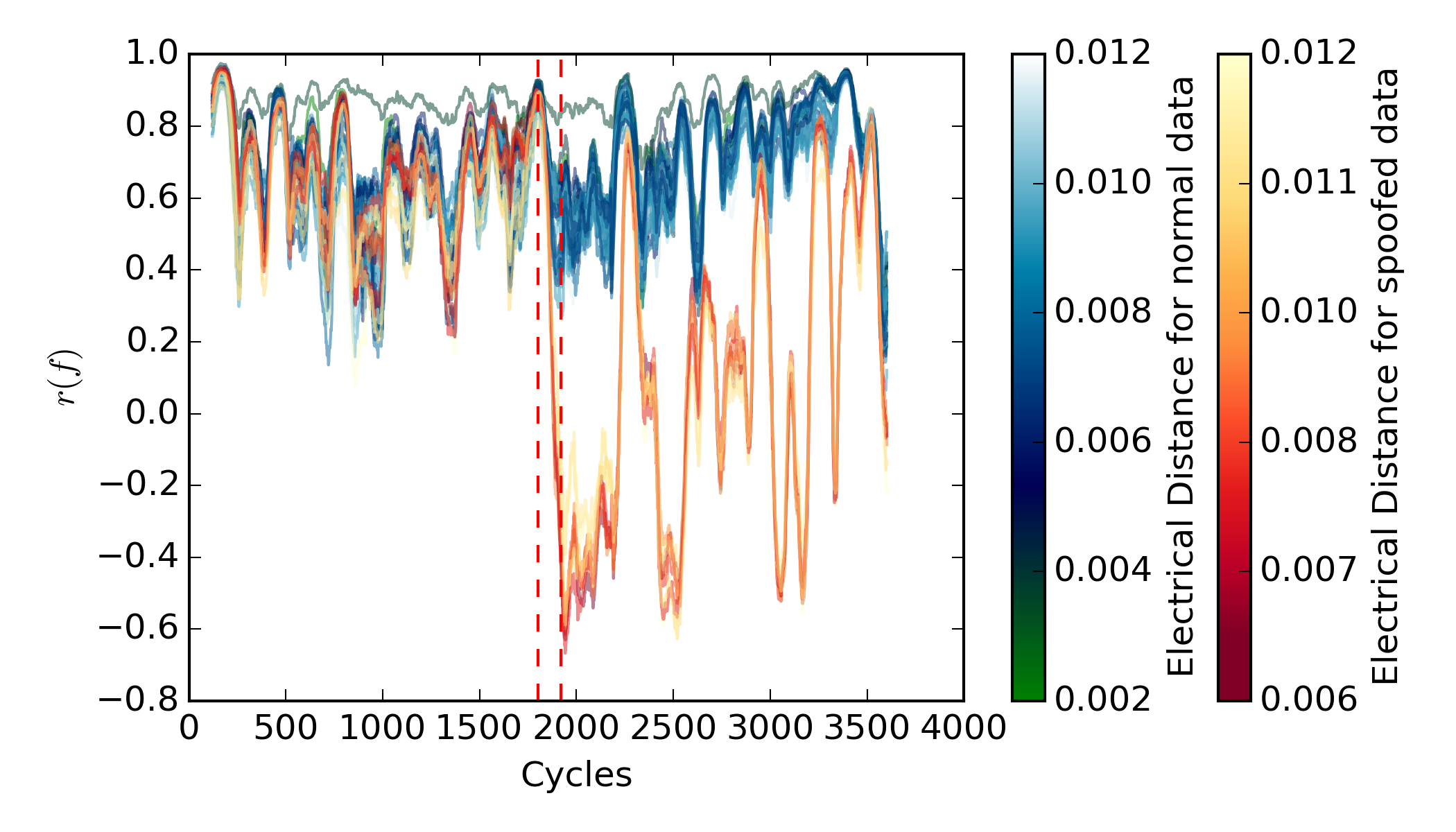}}
    \subfigure[300 cycle (5 seconds) window size]{\includegraphics[width=3.5in]{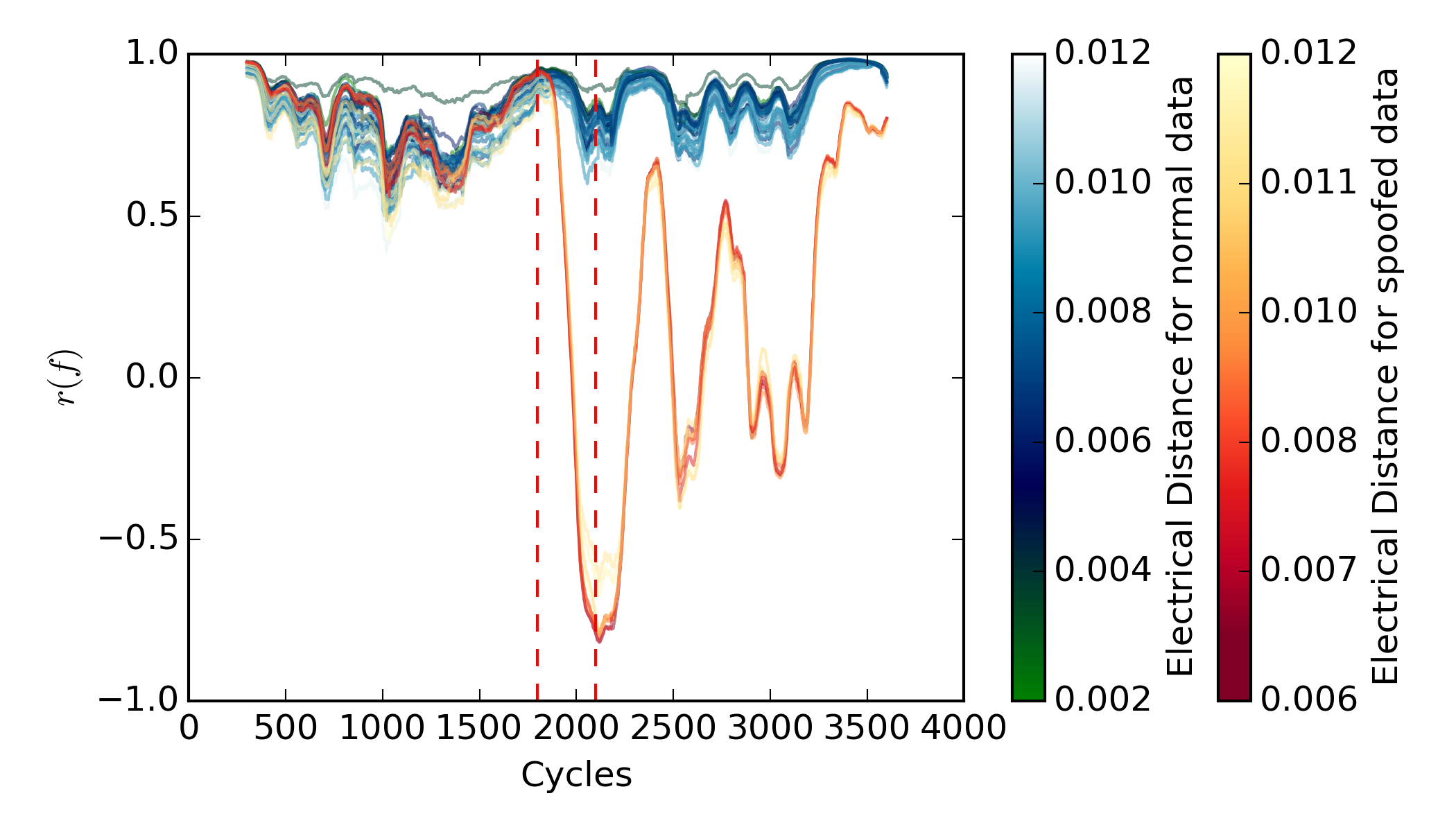}}
\caption{Plot of the Pearson correlation coefficient of frequency $r(f)$ using a window size of (a) 120 cycles and (b) 300 cycles. The spoofed signal affects correlation with nine other PMUs, results in nine deviating correlation plots, shown as yellow-red curves. These deviate markedly from correlation plots between non-spoofed PMUs, blue-green. }
\label{fig:r_freq_corr}
\vspace{-.25in}
\end{figure}

Figure~\ref{fig:r_freq_corr} shows the impact of data injection (S1: Mirroring Spoof) upon the correlation of frequency measurements between PMUs.  Correlation of frequency measurements between all PMUs is greater than 0.5 prior to the spoofing event at $1800$ cycles, as shown by the yellow-red and blue-green gradient curves on the left side of the Figure. The color gradients indicate the electrical distance between each pair of PMUs; PMUs that are electrically close show higher correlation.    

Injecting spoofed data at one PMU affects correlation between that PMU and the nine others. As such, after the spoofing event begins a set of nine curves (those marked by the yellow-red gradient) decouples rapidly from the others. As shown in Figure~\ref{fig:r_freq_corr}, the nine $r(f)$ correlation plots between the target PMU and all others begins to decrease shortly after the data injection begins at 1800 cycles. In order to measure the extent and impact of decorrelation for these signals, we formalize the following two metrics:

\textit{Maximum Correlation Deviation ($MCD$)}: A measure of the maximum difference between the non-spoofed (\textit{nspf}) data (blue-green gradient) and the spoofed (\textit{spf}) data (yellow-red gradient), calculated as an element-wise Euclidean distance:

\begin{equation}
\label{equ:MCD}
MCD = max\left[\sqrt{(\textnormal{nspf} - \textnormal{spf})^{2}}\right]
\end{equation}

\textit{Maximum Correlation Out-Of-Bounds time ($MCOOB$)}: A measure of the amount of time that the spoofed data remains outside of a $\pm10\%$ bound on the non-spoofed data. This is calculated as a summation of the time where the signal satisfies the following inequality: 

\begin{equation}
\label{equ:MCOOB}
MCOOB = \Sigma_{cycles}\left(0.9\times\textnormal{nspf} > \textnormal{spf} > 1.1\times\textnormal{nspf}\right)
\end{equation}

\begin{figure*}[ht] 
  \centering
    \includegraphics[width=0.75\textwidth]{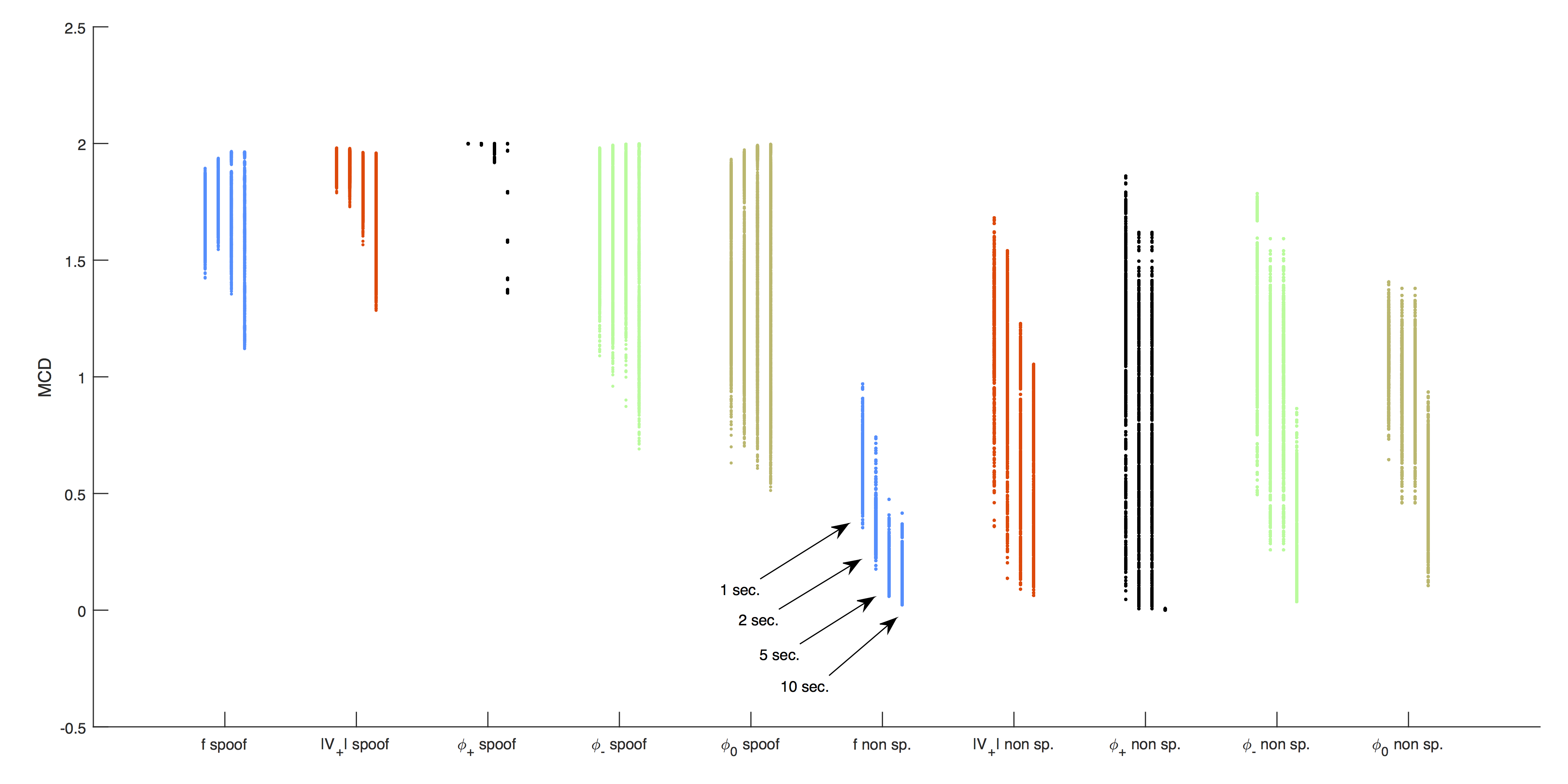}
    \includegraphics[width=0.75\textwidth]{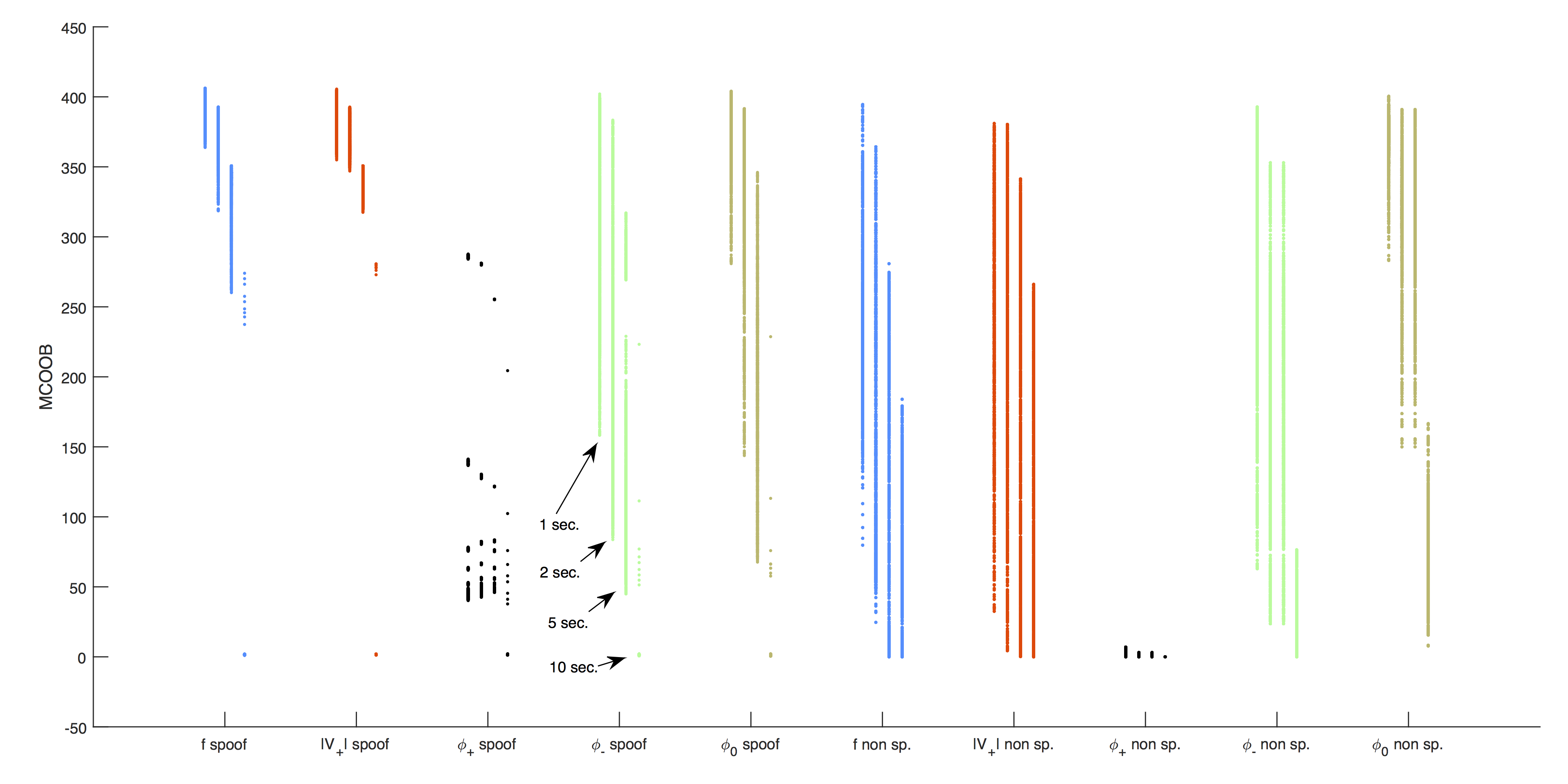}
\caption{Severity of decorrelation indices ($MCD$ and $MCOOB$) across all spoof strategies, correlation windows, and signals of the five feature set ($f$, $|V_{+}|$, $\phi_{+}$, $\phi_{-}$, and $\phi_{0}$). For each panel, the five groups of measures on the left-hand side correspond to spoofed data and the five groups of measures on the right-hand side correspond to non-spoofed data. Within each group, there are four clusters that correspond to the 1, 2, 5, and 10 second window sizes.}
\label{fig:mcd_mcoob}
\vspace{-.25in}
\end{figure*}

From the previous discussion, one can see how the magnitude and timing of decorrelation aren't necessarily coupled homogeneously across signals nor window sizes. Thus, we measured $MCD$ and $MCOOB$ across all available data for our experimental setup (see Figure \ref{fig:mcd_mcoob}). This characterization across magnitude ($MCD$) and time ($MCOOB$) is important to in order to set up and parameterize the spoof detection algorithm presented in Section \ref{sec:SVM}. The distribution of $MCD$ and $MCOOB$ measures in Figure \ref{fig:mcd_mcoob} suggest that for several signals and window sizes one can observe a significant separation between spoofed and non-spoofed data, however, this would not be entirely obvious, for example, for an operator looking at a specific location. 

\section{Spoof Detection via Support Vector Machines}
\label{sec:SVM}

The results presented in Section~\ref{sec:results} show that for the PMUs that are electrically close, the spoofed signals tend to quickly decorrelate from the normal signals, as shown in Figure~\ref{fig:r_freq_corr}. This makes it possible to detect the spoofing attack by monitoring correlations between PMU pairs. However, simply monitoring one signal correlation (such as the frequency correlation $r(f)$) may not guarantee a timely identification of a spoof across all types of spoofs. 
Rather, we expect that robustly identifying spoofed signals requires a careful examination of historical correlation values. Bonneville Power Administration's current installed PMU base contains over 40 devices generating roughly 1.2 TB of data each month.
The challenges associated with storing, accessing, and processing this information in a timely manner will only increase as the installation base expands. Therefore, manual analysis on PMU correlation is not feasible. 

To address these challenges, we use two-class Support Vector Machines (SVM)~\cite{cortes95} to learn a relationship that differentiates spoofed PMU data from normal, untainted, PMU data. A two-class SVM takes as input a set of training examples $\boldsymbol{x}_i, i=1\dots n$, and their associated labels $y_i \in \{-1, 1\}$. In our case, the labels indicate that an example is either normal or spoofed, and each example is defined by an $m$-dimensional vector of features.  For our problem, these features are correlations between signals measured by two PMUs. Based on the previous discussion, we considered two possible features sets. The set of {\em three features} includes positive sequence voltage magnitude and phase-angle correlations ($r_{i,j}\{V{+}\}$, and $r_{i,j}\{\phi_{+}\}$) as well as frequency correlation ($r_{i,j}\{f\}$). The set of {\em five features} adds negative and zero -sequence phase angle correlations ($r_{i,j}\{\phi_{-}\}$ and $r_{i,j}\{\phi_{0}\}$) to the three feature set. All correlations are computed on a trailing 300-cycle window for all pairs of PMUs, $i < j$. 

For a given set of examples, described by their feature vectors $\boldsymbol{x}$, training the support vector machine solves the optimization problem:
\begin{equation}
\begin{split}
 \min_{w,b,\xi} & \frac{1}{2}\boldsymbol{w}^T\boldsymbol{w}+ C \sum_{i=1}^{l}\xi_i \\
 \text{subject to } & y_i(\boldsymbol{w}^T\phi(\boldsymbol{x}_i)+b) \ge 1 - \xi_i , \\
 & \xi_i \ge 0, i = 1,\dots ,l
 \end{split}
\label{eqn:svm}
\end{equation}

Where $\xi_i$ are non-negative slack variables that allow a soft margin (one in which some instances are incorrectly classified). The function $\phi$ transforms the input vector $\boldsymbol{x}_i$ into a higher dimensional space, $C$ is a regularization parameter, and the pair $\boldsymbol{w},b$ defines the hyper-plane that will serve as a classifier between the class labels $\{-1, 1\}$. Equation~\ref{eqn:svm} is easy to interpret, but for efficiency, it is the dual form of this equation that is typically solved. Although not presented here, the dual form makes use of a kernel function, Equation~\ref{eqn:Kxx}, that defines the shape of the decision boundary given a set of support vectors $\boldsymbol{x}_i$. 

\begin{equation}
    K(\boldsymbol{x}_i, \boldsymbol{x}_j) \equiv \phi(\boldsymbol{x}_i)^{T}\phi(\boldsymbol{x}_j)
    \label{eqn:Kxx}
\end{equation}

We leverage the Python library sci-kit learn for a Support Vector Machine implementation based on libsvm\cite{scikitlearn,libsvm}. 

\subsection{Performance Measures}
\label{sec:performance-metrics}

Once trained, the support vector machine will be tested using a new set of labeled data $x'_i$, $y'_i$. Performance on this {\em test set} will be assessed with four metrics:
\begin{itemize}
\item {\bf Sensitivity}: measures the ability to correctly detect spoofed signals and is calculated as the number of true positives (spoofed examples identified as such) divided by the number of total positives (the total number of spoofed examples which is the sum of true positives and false negatives). Sensitivity ranges from $0\%$ to $100\%$ with an ideal classifier measuring $100\%$ sensitivity.

\item {\bf False Discovery Rate:} measures the propensity to spuriously identify a spoof. This value is calculated as the number of false positives (normal examples identified as spoofs) divided by the number of detected spoofs (false positives plus true positives). False Discovery Rate is equivalent to (1-Precision). FDR ranges from $0\%$ to $100\%$; an ideal classifier has $0\%$ FDR.

\item {\bf F1:} measures performance as a single value when classes are not equally prevalent. It is the harmonic mean of Sensitivity and Precision. F1 score ranges from 0.0 to 1.0, higher values are better.

\item {\bf Latency:} measures how long it takes to consistently identify a spoof once it has begun. In this study, we measure latency as the number of cycles after the spoof begins but before the classifier correctly identifies a string of 30 consecutive cycles as spoofed. Lower values are better. Note that while our experiments are performed on archived PMU data, our approach supports use with streaming data with the same latency characteristics described here. 

\end{itemize}

\subsection{Training and Testing Data}
Using the correlation features described above, we constructed a set of examples for each spoof described in Section~\ref{sec:spoof-types} by applying the spoofing procedure to the last 30 seconds of one selected PMU signal on each of 14 different minutes of data. This approach generates roughly $2\cdot 10^6$ examples from the 14 minutes of data and the 45 PMU pairs $i < j$.
Examples are ``Spoofed'' in the last half of each minute if $i$ is the spoofed PMU, and are ``Normal'' otherwise. This approach yields approximately 5 times as many ``Normal'' examples as ``Spoofed'' examples. 
Given the 14 minutes of data, we use 11 minutes (roughly $1.6\cdot10^6$ examples) for training the SVM, and 3 minutes (roughly $4.5\cdot10^5$ examples) for testing. During training, all correlations features are standardized (normalized to 0 mean and standard deviation of 1). The normalization transforms from the training features are saved so they can later be used to transform testing data prior to being classified.


\subsection{Parameter Selection and Training}
During training, we used the RBF kernel parameterized by the scalar value $\gamma$: 

\begin{equation}
K(\boldsymbol{x}_i, \boldsymbol{x}_j) = \exp{(-\gamma||\boldsymbol{x}_i - \boldsymbol{x}_j ||^2)} 
\label{eqn:RBF}
\end{equation}

\noindent We then split the 11 training minutes into two sets (8 and 3 minutes respectively) and performed a grid search over the $C,\gamma$ parameter space by training on the former set and testing on the later. We performed this search for both the {\em three feature} data and the {\em five feature} data using the mirroring spoof (S1). In both cases, we observed high performance (F1 $>$ .95) across a wide range of parameter settings. However, we observed faster training times and marginally improved F1 scores when using five features instead of three. Thus, in subsequent sections, our results are constructed using five feature training/testing data and a RBF-SVM using the parameters $C=1.0$, $\gamma=0.2$. 

Simultaneous to our exploration of feature set size and SVM-parameter settings, we also explored two methods of labeling the training data. Recall that each minute of data contains a spoof in the final 30 seconds and that the Support Vector Machines take, as input, correlations between signals from pairs of different PMUs. Because the correlations are computed on a trailing 300-cycle window, when the spoof begins, the correlation window contains 299 cycles of non-spoofed data, and only 1 cycle of spoofed data. Intuitively, it seems that trying to identify the spoof when the correlation window is dominated by normal data would lead to a substantially higher False Discovery Rate (FDR), an undesirable outcome in a real operating situation. To mitigate this effect, we labeled the {\em training data} as Spoofed when the correlation window contains a majority of spoofed data, and as Normal otherwise (we call this training configuration {\em late timing}). In testing, however, data is labeled as Spoofed when the correlations window contains one or more elements of spoofed data as this is the moment when the spoof actually begins (we call this {\em early timing}).  Thus, our late timing training strategy would be expected to trade off an improvement in False Discovery Rate for a potentially slower latency in recognizing the spoof when compared to an early timing training strategy\footnote{We used the same 8/3 split of the training set to examine impact of training with early vs. late timing in an effort to confirm the intuition described above. Contrary to our expectation, the initial validation showed improved performance in terms of F1, Sensitivity and Latency, with little cost to False Discovery Rate. However, after training on the full 11 minutes, we did observe a severe rise in FDR when testing on the reserved 4 minutes of test data thereby justifying the choice of late timing training.}.

\subsection{Spoof-Specific Classification Results} \label{sec:spoof-specific-results}

Table~\ref{tab:per-spoof-svm} illustrates the performance of our SVM classifiers where each classifier is trained on a distinct type of spoof. For completeness, in the first three columns we show the total count of True Positives (Spoofed data detected as such), False Positives (Normal data detected as a Spoof) and False Negatives (Spoofed data detected as Normal).  From these raw data, we also show summary statistics: Latency, Sensitivity, False Discovery Rate and F1 score described in Section~\ref{sec:performance-metrics}.


The table illustrates that the Spoof-specific classifiers all perform very well identifying over 77\% of the spoofed correlations while maintaining a low False Discovery Rate. Note from a functional perspective, these tests are likely to be overly stringent: a perfect test score can only be obtained by correctly classifying each of the momentary correlations. Higher Sensitivity helps to ensure that the spoof will be detected reasonably early, which, in a real-world setting is likely to be the most salient goal.  

Our latency measurements indicate that all SVMs are able to detect their respective event 
types within 4 seconds (240 cycles), and often much more quickly.  This is particularly impressive given that the correlation windows are 300 cycles long, so a 240-cycle latency indicates the event can be detected even before the correlation window is filled with spoofed data. 

\begin{table*}[ht]
\centering
\begin{tabular}{l|rrrrrrr}
 Spoof Type & True$+$ & False$+$ & False$-$ & Latency & Sensitivity & FDR & F1  \\  
 \hline
 S1; Mirroring & 40749 & 105 & 7905 & [68, 148] & 83.75\% & 0.26\% & 0.911 \\  
 S2; Polynomial & 39194 & 26 & 9460 & [4, 204] & 80.56\% & 0.07\% & 0.892 \\ 
 S3.1; Dilation x2 & 42163 & 301 & 6491 & [120, 209] & 86.66\% & 0.71\% & 0.926\\ 
 S3.2; Dilation x3/2 & 41077 & 793 & 7577 & [144, 223] & 84.43\% & 1.89\% & 0.908\\ 
 S3.3; Dilation x4/3 & 39602 & 870 & 9052 & [154, 226] & 81.40\% & 2.15\% & 0.889 \\ 
 S3.4; Dilation x5/4 & 39666 & 900 & 8988 & [157, 214] & 81.53\% & 2.22\% & 0.889\\ 
 S3.5; Dilation x6/5 & 39054 & 1102 & 9600 & [154, 213] &  80.27\% & 2.74\% & 0.879\\ 
 S3.6; Dilation x7/8 & 38138 & 654 & 10516 & [134, 223] & 78.39\% & 1.69\% & 0.872 \\ 
 S3.7; Dilation x8/9 & 37601 & 520 & 11053 & [130, 217] & 77.28\% & 1.36\% & 0.867 \\ 

\end{tabular}
\caption{SVM Performance When Trained on Specific Spoof Types. Spoof: $+$ Normal: $-$ }
\label{tab:per-spoof-svm}
\vspace{-.25in}
\end{table*}

\begin{figure*}[ht]
\centering
\subfigure[Mirroring (S1): Normalized Frequency Correlation]{%
\includegraphics[width =.9\columnwidth]{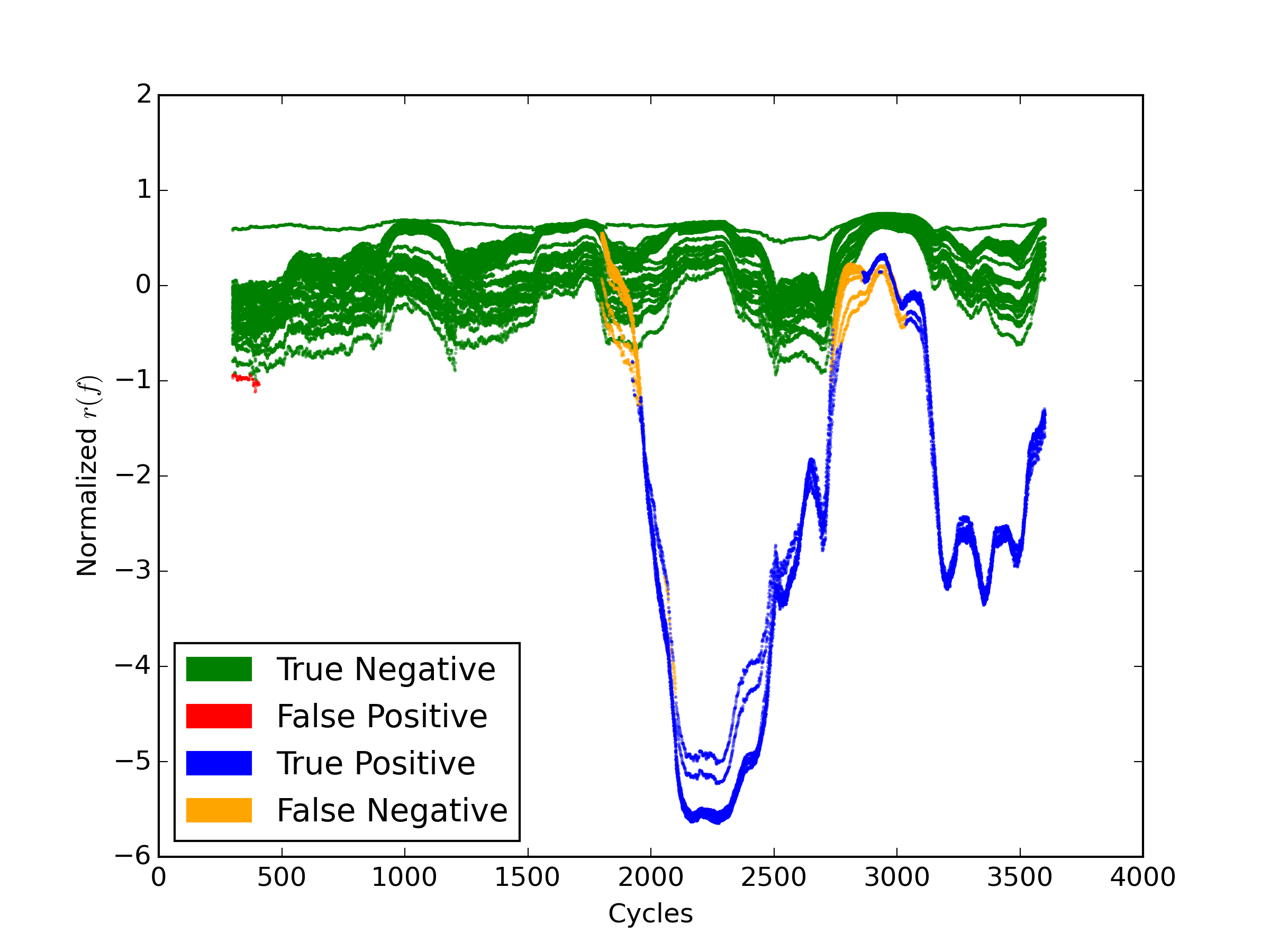}}
\qquad
\subfigure[Time Dilation x8/9 (S3.7): Normalized Frequency Correlation]{%
\includegraphics[width =.9\columnwidth]{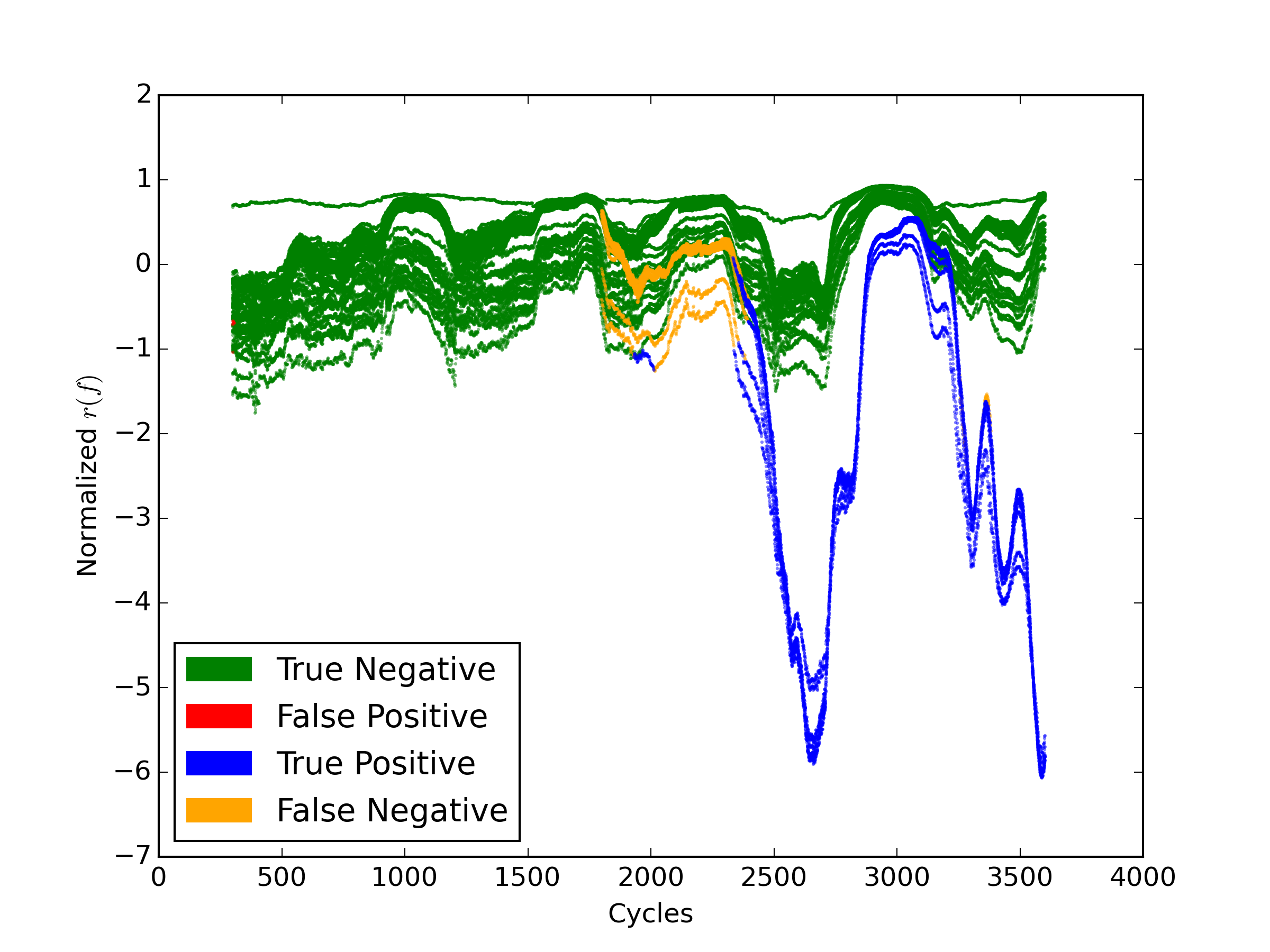}}
\centering




\caption{Frequency Correlations for all pairs of PMUs prior to and during two distinct spoofs.}
\label{fig_sim}
\vspace{-.25in}
\end{figure*}

Figure~\ref{fig_sim} illustrates the classifier's performance on two typical spoofed signals. Figure~\ref{fig_sim}a shows the Mirroring Spoof (S1), while Figure~\ref{fig_sim}b shows a Time Dilation Spoof (S3.7). The green lines toward the top of each plot show correlations between pairs of normally operating PMUs. Recall that for each correlation feature, mean and standard deviations were obtained during training to scale the features prior to learning and classification; thus values are not bounded by $[-1,1]$.  Halfway through the minute, at Cycle 1800, the spoof begins. 

For the Mirroring Spoof (S1), we see relatively rapid decorrelation of the spoofed signal's frequency (yellow and blue lines diverging from the cluster at approximately 1800 cycles). The decorrelation happens much more slowly for the Time Dilation Spoof (S3.7). For both spoofs, there are significant periods during which the frequency correlation values for the spoofed signal are well within the range exhibited for normally operating pairs of PMUs. For example, one such region occurs in both plots between cycles 2700 and 3200. Finally, note that although the frequency correlation in Figure~\ref{fig_sim}b is slow to reach maximum decorrelation, a period of early detection still occurs before the correlation window fills with spoofed data at cycle 2100.

\subsection{Spoof Detection Ensemble}

The results presented in Section~\ref{sec:spoof-specific-results} illustrates that in general our spoof-specific Support Vector Machines can accurately detect spoofed PMU signals if they are trained on one specific target type of spoof. However, given the fact the nature of the spoof attacks may not always be available a priori, a more generalized approach is essential. In Machine Learning, Ensemble Methods refer to a general approach of combining multiple imperfect classifiers to produce a single, more robust, discriminator~\cite{Opitz99popularensemble}. Here in this work, we develop an ensemble which combines the spoof-specific Support Vector Machines and collectively makes decisions on unknown signals, aiming for obtaining better predictive performance than that of individual classifiers. 

In Section~\ref{sec:spoof-specific-results}, we developed nine SVM classifiers, each targeted toward a particular spoof. To assess the generalization ability of these classifiers when used together, we combined eight of the nine SVMs described above and then tested this ensemble on the spoof whose classifier was left out. So, when classifying the Mirroring Spoof (S1), we used an ensemble of classifiers build from the eight SVMs for S2 and S3.1--S3.7. In this fashion, we can observe the ability of the ensemble to identify spoofs that are unknown a priori and compare the ensemble's performance to the performance of a spoof-specific classifier.

When presented with an example to label as spoof or normal data, the ensemble functions by asking each constituent SVM to cast a vote for one label or another. We set a threshold to indicate the minimum number of votes that must be cast for an example to be labeled as a spoof and allow that threshold to range from 1 to 8. Table~\ref{tab:ensemble} illustrates the performance of the ensemble across all threshold values.  The first three columns of the table hold the Sensitivity, False Discovery Rate and F1 Scores respectively. The final three columns show latency ranges for the S1 (Mirroring) Spoof, S2 (Polynomial) Spoof, and the seven S3 (Dilation) Spoofs respectively. For the final column showing latency on the seven S3 spoofs, we show the latency range in square brackets followed by the mean latency because we observed that for many thresholds, the maximum value is determined by an outlier.

Overall, the performance of the ensemble follows intuition: for low thresholds, sensitivity is high, but so is the false discovery rate.  As threshold increases, both sensitivity and false discovery rate drop, and latency tends to increase. Overall performance, as measured by F1 Score, is relatively stable and higher than that reported by most of the spoof-specific classifiers except at extreme thresholds. However, a threshold between 4 and 6 seems to obtain the most reasonable trade-off in performance. Note that the threshold value of 5 corresponds to a majority-rules voting scheme which is a common method for combining classifiers~\cite{ensemblemethods}. Interestingly, across all thresholds, the ensemble's worst-case latency rates for the Mirroring Spoof and Polynomial Spoof (fourth and fifth column) are notably reduced as compared to the per-spoof SVMs presented in Section~\ref{sec:spoof-specific-results}. For the Dilation Spoof (column 6) the best case latency improves while mean latency remains in the latency range experienced by the spoof-specific SVMs except when the threshold is raised close to extreme values (7 or 8), at which average latency experiences a steep rise and maximum latency rises above 500 cycles.

\begin{table}[ht]
\centering
\begin{tabular}{c|rrrrrr}
T & Sensitivity & FDR & F1 & S1L & S2L & S3L \\ \hline
     1 & 90.10\% & 14.84\% & 0.876 & [16, 50] & [2, 42] & [44, 203] 103\\
     2 & 88.74\% & 5.97\% & 0.917 & [21, 58] & [2, 43] & [67, 208] 135\\
     3 & 87.68\% & 3.42\% & 0.919 & [26, 87] & [3, 45] & [91, 215] 155\\
     4 & 86.87\% & 1.71\% & 0.922 & [26, 89] & [3, 47] & [95, 297] 177\\
     5 & 85.96\% & 1.10\% & 0.920 & [28, 90] & [4, 51] & [107, 464] 192\\
     6 & 85.07\% & 0.63\% & 0.917 & [30, 92] & [5, 52] & [116, 474] 203\\
     7 & 82.69\% & 0.20\% & 0.904 & [32, 95] & [4, 62] & [120, 512] 241\\
     8 & 52.48\% & 0.22\% & 0.688 & [33, 114] & [7, 111] & [120, 534] 275\\
\end{tabular}
\caption{Performance of the Classifier Ensemble With Given Threshold (T) for Spoof Detection}
\label{tab:ensemble}
\end{table}





\section{Conclusion} \label{sec:conclusion}
We have proposed a novel approach for addressing the challenge of detecting spoofed data from phasor measurement units (PMUs).  Our approach uses machine learning techniques and correlation coefficients between measurement parameters of electrically close PMUs and, because of the short latency times for detection, is capable of supporting streaming data and real-time spoof detection in a live setting.  Two-class support vector machines (SVMs) are trained using both normal and spoofed PMU data. Nine spoofed data models are derived using three different spoofing playback schemes. The SVMs are evaluated using PMU data collected from Bonneville Power Administration's extensive PMU network. Experimental results show the SVMs can effectively detect spoofed signals; against the nine spoof data sets, results show high sensitivities and F1 scores while concurrently demonstrating low false discovery rates and reasonable latencies. Detection is enhanced when using a majority-rules ensemble discriminator comprised of SVMs trained on the nine spoof data sets.  For future work, we are investigating the effectiveness of other machine learning methods in addressing this problem, such as one-class learning and on-line learning methods. 

\section*{Acknowledgments} 

This research is supported by student scholarships from Oregon BEST and the Bonneville Power Administration. We thank both Portland General Electric and the Bonneville Power Administration for providing PMU data.

\bibliographystyle{ieeetr}

\bibliography{bibfile.bib}

\begin{IEEEbiographynophoto}
{Rich Meier} is completing his Master's in Electrical and Computer Engineering at Oregon State University.
\vspace{-4em}
\end{IEEEbiographynophoto}

\begin{IEEEbiographynophoto}
{Jordan Landford} is pursuing his M.S. in Electrical and Computer Engineering at Portland State University.
\vspace{-4em}
\end{IEEEbiographynophoto}

\begin{IEEEbiographynophoto}
{Robert Bass} received the Ph.D. degree in electrical engineering in 2004 from The University of Virginia. 
\vspace{-4em}
\end{IEEEbiographynophoto}

\begin{IEEEbiographynophoto}
{Eduardo Cotilla-Sanchez} received the Ph.D. degree in electrical engineering in 2012 from The University of Vermont.  
\vspace{-4em}
\end{IEEEbiographynophoto}

\begin{IEEEbiographynophoto}
{Scott Wallace} received the Ph.D. degree in computer science and engineering in 2004 from The University of Michigan, Ann Arbor.
\vspace{-4em}
\end{IEEEbiographynophoto}

\begin{IEEEbiographynophoto}
{Xinghui Zhao} received the Ph.D. degree in computer science in 2012 from The University of Saskatchewan. 
\vspace{-4em}
\end{IEEEbiographynophoto}

\begin{IEEEbiographynophoto}
{Richard Barella} is pursuing his B.S. in Computer Science at Washington State University, Vancouver.
\vspace{-4em}
\end{IEEEbiographynophoto}

\vfill

\end{document}